# Inducing Constraint Grammars

Christer Samuelsson[1], Pasi Tapanainen[2] and Atro Voutilainen[2]

[1] Universität des Saarlandes, FR 8.7, Computerlinguistik
Postfach 1150, D-66041 Saarbrücken, Germany
christer@coli.uni-sb.de
[2] Research Unit for Multilingual Language Technology
P.O. Box 4, FIN-00014 University of Helsinki, Finland
{tapanain,avoutila}@ling.helsinki.fi

**Abstract.** Constraint Grammar rules are induced from corpora. A simple scheme based on local information, i.e., on lexical biases and next-neighbour contexts, extended through the use of barriers, reached 87.3 % precision (1.12 tags/word) at 98.2 % recall. The results compare favourably with other methods that are used for similar tasks although they are by no means as good as the results achieved using the original hand-written rules developed over several years time.

## 1 Introduction

The present article describes experiments with inducing Constraint Grammars from annotated corpora. As described in Section 2, Constraint Grammar is a rule-based framework for morphological disambiguation and shallow syntactic parsing, where the rules are hand-coded by a linguistic expert. The present work does not aim at replacing the human grammar developer, but at supporting the grammar development task. It enables creating a first version of the grammar, which the grammarian can enhance in various ways, e.g. by discarding rules that are obviously incorrect, by adding additional constraints to rules that overgeneralise, and by adding linguistically motivated rules to cover phenomena that cannot readily be inferred from data. The only real advantage that the system has over the human is the ability to quantify what phenomena are common and what are not. Knowledge of this is essential for efficient grammar development, and the system can thus also find disambiguation rules that the human has overlooked.

The remainder of the article is organised as follows: The Constraint Grammar framework is presented in Section 2, while Section 3 describes the details of the various formats of the induced grammar rules. The learning procedure is explained in detail in Section 4, and the experimental results are reported and discussed in Section 5.

## 2 Constraint Grammar Framework

Constraint Grammar (CG), originally proposed by Karlsson [3], and fully documented in Karlsson et al. [4] and Tapanainen [6], is a reductionistic parsing framework based on the introduction and subsequent resolution of morphological and shallow syntactic ambiguities. The first mature CG parser, the English CG parser EngCG [11], consists of the following sequentially applied modules:

1. Tokenisation
2. Lookup of morphological tags
   (a) Lexical component
   (b) Rule-based guesser for unknown words
3. Resolution of morphological ambiguities
4. Lookup of syntactic function tags
5. Resolution of syntactic ambiguities

EngCG uses a morphological analyser, EngTWOL, with 90 000 lexical entries and a morphological description with about 180 ambiguity-forming morphological readings. Words not represented in EngTWOL are analysed with an accurate rule-based guesser. The following is an example of the output from EngTWOL:

```
"<campaign>"
        "campaign" <SV> <P/for> V SUBJUNCTIVE VFIN
        "campaign" <SV> <P/for> V IMP VFIN
        "campaign" <SV> <P/for> V INF
        "campaign" <SV> <P/for> V PRES -SG3 VFIN
        "campaign" N NOM SG
```

It contains the various readings of the word "campaign". This should be understood as follows: The word form is "campaign" as indicated by "<campaign>". There are five different readings. The word stem is "campaign" in all five of them as indicated by "campaign". The first four are verb readings, which is indicated by the V feature, while the last one is a noun reading bearing the N feature. The verb readings are in turn subjunctive ("They insisted that she campaign more effectively."), imperative ("Campaign more effectively!"), infinitive ("It is important to campaign more effectively."), and present indicative ("We campaign more effectively."). The first two readings, and the fourth one, are finite verb forms (VFIN). The first two features <SV> and <P/for> pertain to the possible syntactic subcategorization patterns of the verb readings: "They campaign." (intransitive) and "They campaign for it." (prepositional-phrase complement with "for" as the preposition).

The disambiguator uses a grammar of 1 200 constraint rules that refer to the global context and discard illegitimate morphological analyses in contexts specified by local or global contextual conditions. There are also some 250 heuristic rules for resolving remaining ambiguities.

EngCG is reported [11, 9, 7, 8] to assign a correct analysis to about 99.7 % of all words; on the other hand, each word retains on average 1.04–1.09 alternative analyses, i.e. some of the ambiguities remain unresolved. If also the heuristic constraints are used, about 50 % of the remaining ambiguities are resolved, but the error rate goes up to about 0.5 %.

# 3 Rule Typology

This section describes the different types of rules induced in the experiments.

## 3.1 Basic local-context rules

The basic format of the induced rules is:

REMOVE (V) (-1C (DET));

This should be read as: "Discard (REMOVE) any reading with the verb feature (V) if *all* readings (C) of the preceding word (-1) have the determiner feature (DET)." Omitting the C in -1C would mean "if *any* reading of the preceding word ..." The underlying idea of this particular rule is that readings bearing the V feature cannot intervene between a determiner and the head of the noun phrase. One might object that this would not hold for participles such as "given" in "the given example", but in the Constraint Grammar framework, these readings do not bear the V feature. The converse rule would be

REMOVE (DET) (1C (V));

and should be read as: "Discard any reading with the determiner feature if all readings of the *following* word have the verb feature".

These basic local-context rules can be induced by automatically inspecting an annotated corpus, and noting what features do not, or very seldom, occur on neighbouring words. In fact, the interesting quantities are the bigram feature counts and, as a comparison, the unigram feature counts, as explained in Section 4.1.

## 3.2 Combined local-context rules

Another well-motivated rule would be

REMOVE (V) (-1C (PREP));

and a number of other rules that discard verb readings if the preceding word carries some particular feature. These features can be collected into sets that can be referred to in the rules:

REMOVE (V) (-1C SET1);

Here SET1 is a set of features, and if all readings of the preceding word has some feature in this set, we can discard any reading of the current word that has a verb feature. In the example, SET1 would consist of DET and PREP, and other features as appropriate.

Note that this gives us more disambiguating power than the two original rules together; if the preceding word has one determiner reading and one prepositional reading, neither of the two original rules would be applicable, while the latter, combined local-context rule, would be. These rules can be automatically constructed from the previously discussed basic local-context rules.

### 3.3 Barrier rules

Barriers rules enable reference to tags whose precise position, relative to the ambiguous word in Position 0, is not known. In a barrier rule, some context conditions contain two parts each: (i) one part identifies some tag(s) somewhere to the left or right; (ii) the other (the barrier) states, what features are not allowed to occur in an intervening position. For instance, the following rule removes all readings with the tag V, if somewhere to the left there is an unambiguous determiner DET and there are no members of the set NPHEAD to the right of it, up to position -1.

REMOVE (V) (*-1C (DET) BARRIER NPHEAD);

The star * in (*-1C) means "one or more words", so *-1C (DET) means that for some word to the left, it is the case that all readings have the determiner feature DET. BARRIER is a reserved word of the CG description language and NPHEAD is a set of features that the grammarian has postulated, just like the set SET1 above. NPHEAD is here taken to be the set of features of the words that can function as heads of noun phrases, e.g., N for nouns, PRON for pronouns, NUM for numerals, etc. BARRIER means that there are no intervening words with any reading with any feature in the set following it. Thus, (*-1C (DET) BARRIER NPHEAD) means that somewhere to the left, we have a word that must bear the DET feature, and between this word and the current one, there are no words with remaining readings with any feature in the set NPHEAD, i.e., that can function as the head of the noun phrase. The intuition of this barrier rule is thus that if we have seen a verified determiner to the left, but no candidate NP head after that, we can safely remove all verb readings.

These rules can be induced from the basic local-context rules by noting what features actually occur between the features specified in those rules, e.g., by noting what features occur between determiners and verbs. These are collected and form the barrier sets, as described in Section 4.3.

### 3.4 Lexical rules

A third type of rule concerns rare readings of particular words, for example the verb reading of "table" as in "table the motion". The idea here is to see how many times a particular feature is proposed for a certain word in proportion to how many times it is actually in the correct reading. If this feature is not very often in the correct reading, it might be a good idea to remove any readings containing it. This would be effected by the rule

REMOVE (V) (0 ("<table>") );

The zero (0) refers to the current word and "<table>" refers to the word form "table". This rule will unconditionally remove the verb reading of the word form "table". It may seem a bit strange to first propose a particular reading for a word in the morphological analyser, and then write a rule that directly allows the disambiguator to discards it, but there is a very good reason for this: *The disambiguator is not allowed to remove the last remaining reading!* Thus, the system employs a Sherlock-Holmes strategy; if other rules have eliminated all other possible readings, then the remaining one, however unlikely, is the true one.

### 3.5 Rare-feature rules

Similarly, features that are very rarely the correct one, independent of what word form they are assigned to, can be removed in the same way. For example, the subjunctive reading of verbs is not often the correct one. The following rule discards these subjunctive readings:

```
REMOVE (SUBJUNCTIVE);
```

The last two rule types utilise the fact that it is possible to stratify the set of grammar rules, so that the disambiguation is first carried out with a first set of rules until no further readings can be eliminated, then with the first and a second set of rules, etc.

## 4 Learning Strategy

In this section, we describe how the various types of rules can be induced.

### 4.1 Local-context rules

First, unigram and bigram feature statistics are collected. Basic local-context rules such as

```
REMOVE (FEATURE) (-1C (CONTEXT));
REMOVE (FEATURE) (1C (CONTEXT));
```

remove any readings of the current word containing the feature **FEATURE** if all readings of the previous (or next) word contain the feature **CONTEXT**. These rules are induced if the probability of **FEATURE** drops drastically when conditioned on **CONTEXT**, i.e., if:

$$\frac{P(\text{FEATURE} \mid \text{CONTEXT})}{P(\text{FEATURE})} < \text{Threshold}$$

Note that this probability ratio is related to the mutual information statistics of **FEATURE** and **CONTEXT**, see [5], Section 2.2.2, and we will refer to this quantity as the score of the rule. Note also that due to the fact that each correct reading of any word can have a number of features, the probabilities do not necessarily sum to one over the features. P(**FEATURE** | **CONTEXT**) should therefore be interpreted as the probability of **FEATURE** showing up in the correct reading given **CONTEXT**.

Two modifications were made to this to avoid problems with sparse data. Firstly, only features and contexts with a reasonably high frequency count are allowed to participate in this phase. In the actual experiments, they were required to be at least 100. Secondly, instead of estimating P(FEATURE | CONTEXT) directly from the relative frequency of the feature in the context, a 97.5 % upper limit $\tilde{P}$ of this quantity is calculated. If there are no observations of the feature in the context, and if the frequency count of the context is $N$, this will be

$$\tilde{P} = 1 - \sqrt[N]{0.025} \tag{1}$$

Otherwise, with a non-zero relative frequency $f$, the usual (extended Moivre-Laplace) approximation using the normal distribution is employed (see, e.g., [5], Section 1.6):

$$\tilde{P} = f + 1.96 \cdot \sqrt{\frac{f \cdot (1-f)}{N}} \tag{2}$$

Seeing that $N$ was at least 100, this is an acceptable approximation.

Basic local-context rules with the same effect, and referring to the same neighbour (i.e., to the left or to the right), are collapsed into combined local-context rules with more disambiguating power as discussed in Section 3.2.

### 4.2 Rare-reading rules

Lexical rules are of the form

   REMOVE (FEATURE) (0 (WORD) );

and discard the feature FEATURE of the word form WORD. They are induced if

$$\frac{\text{P(FEATURE | WORD)}}{\bar{P}_{FW}} < \text{Threshold}$$

where $\bar{P}_{FW}$ is the average probability over all features and words. Also here, an upper bound for P(FEATURE | WORD) is used instead of using this probability directly, and this bound is established exactly as in the previous case.

Similarly, rare-feature rules of the form

   REMOVE (FEATURE);

unconditionally discard FEATURE regardless of which word bears it, and they are induced if

$$\frac{\text{P(FEATURE)}}{\bar{P}_F} < \text{Threshold}$$

where $\bar{P}_F$ is the average probability over all features. Again, an upper bound is used instead of the probability itself.

## 4.3 Barrier rules

Barriers are established by collecting sets of candidate barrier features from the training data. One such set is constructed for each occurrence of two features that are ruled out as neighbours by any basic local-context rule. The candidate barrier set then simply consists of all features occurring between the two features. From the collection of candidate barrier sets, a minimal set of separating features is constructed for each feature pair using weighted abduction.

For example, if the only observed sequences of DET...V are

DET ADJ N PCP2 V (as in "The huge costs incurred are...")
DET NUM V (as in "The two will...")
DET N ADV V (as in "The shipments often arrive...")

we construct the candidate barrier sets

{ADJ,N,PCP2},{NUM} and {N,ADV}

Assuming that N is in the barrier set explains the first and third example, and assuming that NUM is in the barrier set explains the second one. It is easy to verify that no other barrier set of size two or less explains the observed sequences, and {N,NUM} is therefore chosen as the final set of barriers.

Here weighted abduction essentially means that we must choose (at least) one feature in each candidate barrier set. The cost of selecting a feature that has not previously been selected from any candidate barrier set is one unit, while the features that have already been selected from some candidate barrier set may be reused free of charge.

More formally, a Horn-clause program is constructed where each example will result in one clause for each candidate barrier feature. The conjunction of the examples is then proven at minimal cost. The examples above will result in the program

$$\begin{array}{ll} Ex_1 \leftarrow ADJ \quad (1) & Ex_2 \leftarrow NUM \quad (4) \\ Ex_1 \leftarrow N \quad\quad (2) & Ex_3 \leftarrow N \quad\quad (5) \\ Ex_1 \leftarrow PCP2 \quad (3) & Ex_3 \leftarrow ADV \quad (6) \end{array}$$

and the goal G to prove is $Ex_1$ & $Ex_2$ & $Ex_3$. Any RHS literal, i.e., any feature, may be assumed at the cost of one unit. We prove the goal G by employing an iterative deepening strategy, i.e., a proof of G is sought first at cost zero, then at cost one, then at cost two, etc. In the example, assuming N and NUM, at a total cost of two units, allows proving G through clauses (2), (4) and (5).

A couple of optimisations can be employed: Firstly, if the intersection of the candidate barrier sets is non-empty, any feature in the intersection can be chosen as a singleton barrier set. In practice, the intersection itself was used as a barrier. Secondly, each singleton candidate barrier set, such as {NUM} above, must be a subset of the final barrier set. This observation allows starting the abduction process from the union of all singleton sets, rather than from the empty set. Despite these optimizations, this turned out to be the most time-consuming phase of the induction process, due to the combinatorial nature of the abduction procedure.

This enables extending each basic local-context rule to long-distance dependencies, limited only by the corresponding induced barrier set. Note that this type of rules gives the learned grammar more expressive power than the rules induced in Brill's [1] learning framework. Also, the way the rules are applied is fundamentally different.

### 4.4 Redundancy and stratification

Some features always co-occur with others (within a reading), in which case there is a risk of inducing redundant rules. For example, the VFIN feature implies the presence of the V feature. Thus, there is no point in having a rule of the form

   REMOVE (VFIN) (-1C (DET));

if there is already a rule of the form

   REMOVE (V) (-1C (DET));

This is dealt with by keeping track of the observed feature co-occurrences and discarding candidate rules that are subsumed by other rules.

In the learning phase, the threshold is varied to stratify the rules. During disambiguation, several rule levels are employed. This means that the most reliable rules, i.e., those extracted using the lowest threshold, and that thus have the lowest scores, are applied first. When no further disambiguation is possible using these rules, the set of rules corresponding to the second lowest threshold is added, and disambiguation continues using these two sets of rules, etc. In the experiments reported in Section 5, ten rule levels were employed.

The threshold values and the subsumption test interact in a non-trivial way; low-score rules subsumed by high-score rules should not necessarily be eliminated. This is dealt with in a two-pass manner: In a first database-maintenance step, rules are only discarded if they are subsumed by another rule with a lower score. In a second step, when constructing each grammar level, redundancy within the upper and lower threshold values is eliminated.

Note that redundancy is more of a practical problem when inducing grammar rules, due to the limitations in available storage and processing time, than a theoretical problem during disambiguation: Exactly which rule is used to discard a particular reading is of no great interest. Also, the CG parser is sufficiently fast to cope with the slight overhead introduced by the redundancies.

## 5 Experiments

A grammar was induced from a hand-disambiguated text of approximately 55 000 words comprising various genres, and it was tested on a fresh hand-disambiguated corpus of some 10 000 words.

The training corpus as well as the benchmark corpus against which the system's output was evaluated was created by first applying the preprocessor and morphological analyser to the test text. This morphologically analysed ambiguous text was then independently disambiguated by two linguists whose task also was to detect any errors potentially produced by the previously applied components. They worked independently, consulting written documentation of the grammatical representation when necessary. Then these manually disambiguated versions were automatically compared. At this stage, about 99.3 % of all analyses were identical. When the differences were collectively examined, it was agreed that virtually all were due to clerical mistakes. One of these two corpus versions was modified to represent the consensus, and these "consensus corpora" were used, one for grammar induction and the other for testing the induced grammar. (For more details about a similar annotation experiment, see [10].)

A reasonable threshold value was established from the training corpus alone and used to extract the final learned grammar. It consisted of in total 625 rules distributed fairly evenly between the ten grammar levels. Of the learned rules, 444 were combined local-context rules, 164 were barrier rules, 10 were lexical rules and 7 were rare-feature rules.

The grammar was evaluated on a separate corpus of 9 795 words from the Brown corpus, manually annotated using the EngCG annotation scheme as described above. There were 7 888 spurious readings in addition to the 9 795 correct ones. The learned grammar removed 6 664 readings, including 175 correct ones, yielding a recall of $98.2 \pm 0.3$ % (with 95 % confidence degree) and a precision of $87.3 \pm 0.7$ %. This result is better than the results reported for Brill's [2] N-best tagger. He reports 98.4 % recall when the words have 1.19 tags on average (corresponding to 82.7 % precision) while the induced Constraint Grammar in the current experiments leaves less readings (1.12 per word) for the equivalent recall. However, the comparison to Brill's figures is only meant as an indication of the potential of our approach; more conclusive comparisons would require (i) accounting for the differences between the tag sets and (ii) the use of larger and more varied test corpora.

When these figures are compared with the reported EngCG performance using a hand-crafted grammar, it is obvious that although the proposed method is very promising, much still remains to be done. However, it should be remembered that this grammar was developed and debugged over several years. Thus, the rôle of the proposed method can be seen in three ways: (1) it is a bootstrapping technique for the development of a new grammar, (2) the remaining ambiguities of a linguistic (hand-written) grammar may be resolved by the empirical information (related work has been done in [7]), or (3) automatic induction may help the grammarian to discover new rules semi-automatically, so that the grammarian can remove the rules that are obviously incorrect and also fix and add sets and further contextual tests to the rules. In general, the exceptions to the rules are hard to detect and accommodate automatically, but using linguistic knowledge, the rules can be fixed relatively easily.

An advantage of the proposed approach is that the formalism itself does not restrict the scope of the rules to, say, bigrams. In the future, the result may be improved, for example, by adding linguistically sound predefined sets to guide the learning process towards better rules. Those sets may also be used to reduce the search space in the learning process, and that may make it possible to increase the number of the contextual tests in the rules to make them more accurate. Generally, the rôles of the different approaches can be characterized as follows: "linguistic knowledge is good for making generalisations, but the discovered rules can better distinguish between what is common and what is not."